\documentclass{llncs}
\usepackage{makeidx}  
\usepackage[]{algorithm2e} 
\usepackage[dvipsnames]{xcolor}
\usepackage{amsmath}
\usepackage{amssymb}
\usepackage{booktabs}
\usepackage{color}
\usepackage{graphicx}
\usepackage{todonotes}
\usepackage{subfig}
\usepackage[utf8x]{inputenc}

\begin{document}
%
%
\title{Extracting textual overlays from social media videos using neural networks}
\titlerunning{Extracting...}  
%


\author{Adam Słucki\inst{1, \, 3}, Tomasz Trzciński\inst{2, \, 3}, Adam Bielski\inst{3},  Paweł Cyrta\inst{3}}

\institute{Polish-Japanese Academy of Information Technology, Warsaw, Poland\\
\and
Warsaw University of Technology, Warsaw, Poland\\
\and 
Tooploox, Poland \\
\email{firstname.lastname@tooploox.com}
}

\authorrunning{A. Słucki, T. Trzciński et al.} 

\maketitle              

\vspace{-0.4cm}
\begin{abstract}
Textual overlays are often used in social media videos as people who watch them without the sound would otherwise miss essential information conveyed in the audio stream. This is why extraction of those overlays can serve as an important meta-data source, e.g. for content classification or retrieval tasks. In this work, we present a robust method for extracting textual overlays from videos that builds up on multiple neural network architectures. The proposed solution relies on several processing steps: keyframe extraction, text detection and text recognition. The main component of our system, i.e. the text recognition module, is inspired by a convolutional recurrent neural network architecture and we improve its performance using synthetically generated dataset of over 600,000 images with text prepared by authors specifically for this task. We also develop a filtering method that reduces the amount of overlapping text phrases using Levenshtein distance and further boosts system's performance. The final accuracy of our solution reaches over 80\% and is au pair with state-of-the-art methods. 
\end{abstract}

\section{Introduction}
\begin{figure}[t!]
    \centering
    \subfloat[Standard frame with text]{{\includegraphics[width=0.4\textwidth]{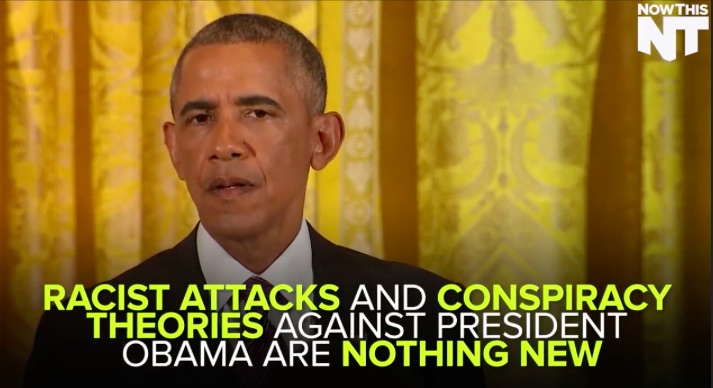} }}%
\qquad
    \subfloat[Superimposed subtitles]{{\includegraphics[width=0.4\textwidth]{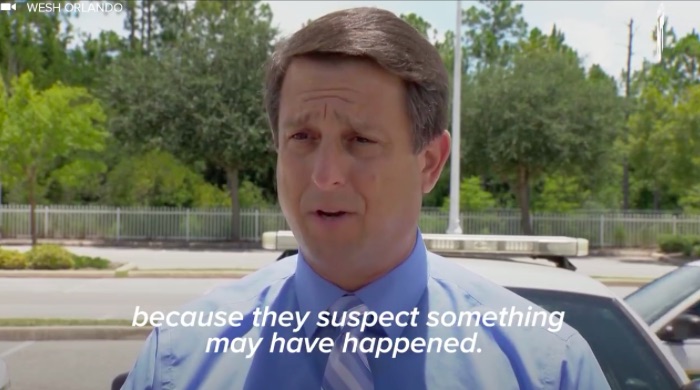} }}%
\qquad  
        \subfloat[Multiple text blocks with various fonts and sizes]{{\includegraphics[width=0.4\textwidth]{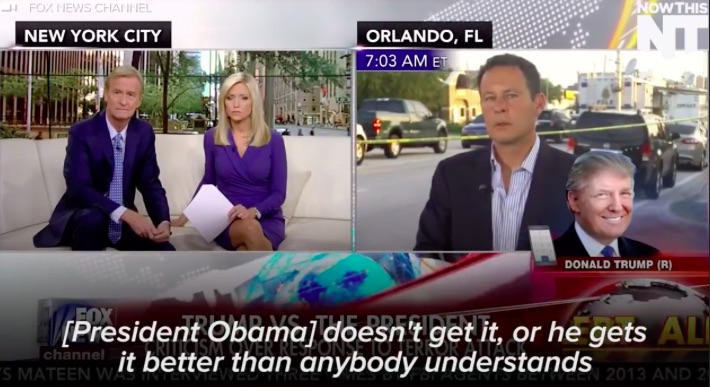} }}%
\qquad
        \subfloat[Text displayed as a part of a background]{{\includegraphics[width=0.4\textwidth]{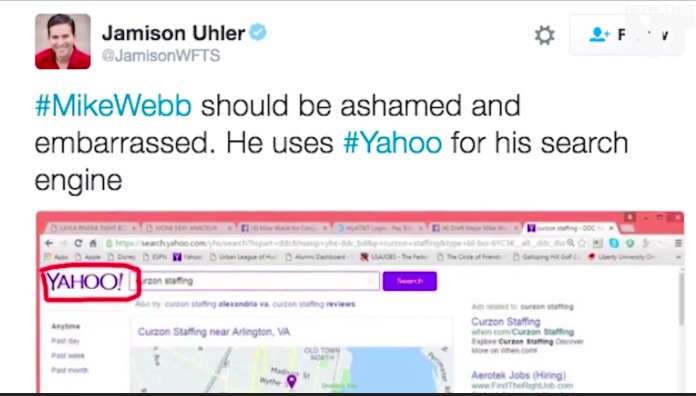} }}%
        
    \caption{Sample frames extracted from social media videos with textual overlays displayed in challenging conditions.}%
    \label{fig:sample_frames}%
\end{figure}

\vspace{-0.4cm} 
Videos published on social media are commonly described only with their title, short summary and unstructured keywords. Extracting additional information from textual overlays such as captions, key ideas or scene level summaries can be a crucial component of a content retrieval system, video classifier or intelligent advertisement targeting. The problem of extracting this information is twofold. First part of the problem is choosing frames on which OCR will be performed and the second is text detection and recognition on those frames. There are many domain-specific difficulties related to the detection and recognition in social media videos. Backgrounds of textual overlays in those videos are rarely solid and contrastive. They are often displayed as part of a background and have various font sizes, colors and combinations. We present examples of frames with text appearing in social media videos in Fig.~\ref{fig:sample_frames}. 




In this paper, we present an entire working pipeline for text extraction tailored specifically for social media videos. We propose a multi-component system that consists of frame extraction, text detection, text recognition and post-processing by text merging and rectification. Fig.~\ref{fig:system} shows an overview of our system. We also propose a method for generating synthetic training data designed for textual overlays commonly appearing in social media videos. Extending our training dataset with the synthetically generated data allows our text recognition model to reduce a word level recognition error by 20\% compared to a general, pre-trained CRNN~\cite{DBLP:journals/corr/ShiBY15} model for text recognition. The main contribution of this work is a complete system that allows its user to extract video overlays with minimal amount of textual overlap and state-of-the-art text detection and recognition results. 
We also describe the details of a training data generation algorithm that takes into account visual characteristics of overlays in online videos and we show how using this algorithm improves the accuracy of our system. Finally, we propose a new method based on the Levenshtein distance that allows to filter out the text appearing in multiple frames and extract the most relevant information presented in a video.

The remainder of this paper is organized in the following manner: In Sec.~\ref{sec:related}, we discuss related works. Sec.~\ref{sec:system} presents our system and in Sec.~\ref{sec:results} we evaluate its performance against baselines. We conclude the paper in Sec.~\ref{sec:conclusions}.

\vspace{-0.2cm} 
\section{Related Work}
\label{sec:related}
\vspace{-0.3cm} 
A significant amount of research focused on addressing the problem of text detection and recognition in images~\cite{DBLP:journals/corr/ShiBY15,DBLP:journals/corr/YaoWZZZCY15,DBLP:journals/corr/TianPHLYT16,Jaderberg14c}. The work \cite{DBLP:journals/corr/YaoWZZZCY15} presents a fully convolutional neural network that is trained for pixelwise classification of text regions in natural scene images. Text recognition is performed with a CRNN model inspired by \cite{DBLP:journals/corr/ShiBY15}, which is further improved through a dictionary based correction. In our system, we also rely on a spellcheck dictionary-based functionality to improve the output of our system. Another approach to text detection in images is presented in~\cite{DBLP:journals/corr/TianPHLYT16}, where the authors use a fast cascade boosting technique to detect single characters. The characters are then merged into lines with min-cost flow network in a post-processing step, similarly to our rectification module. Although the selection of text detection and recognition methods presented above is far from complete, in this work we focus on extracting textual overlays from videos, not images and we review below related works that address this exact problem.  

Detecting and recognizing blocks of text in videos has also gained significant attention from the research community~\cite{Sato-1998-14580,Yang:2016:SRV:2964284.2973811,DBLP:journals/corr/KannaoG16}. In~\cite{Sato-1998-14580}, Sato {\it et al.} present an approach based on extracting and classifying hand-crafted features using a computer vision method. They rely on specific properties of letters to detect blocks of text, segment it into characters and recognize them individually using template matching. To enhance the quality of the input, they leverage temporal consistency of the text blocks displayed across several frames of video using time-based minimum pixels value search. Although fairly effective for videos with high contrast, where the color of textual characters is significantly different then the background, the main limitation of the method is lack of robustness against less contrastive frames. As presented in Fig.~\ref{fig:sample_frames}, this is often not the case for social media videos where various font colors are used and the contrast against the background cannot be guaranteed.

A recent work~\cite{Yang:2016:SRV:2964284.2973811} presents a method that, similarly to~\cite{Sato-1998-14580}, uses a computationally efficient text detection method, in this case the maximally stable external regions (MSER)~\cite{Donoser06}, to generate a set of candidate regions. The regions are then filtered using a binary classifier based on a convolutional neural network architecture and text recognition is done using a similar neural network model. 
The system is capable of providing real-time OCR recognition in videos. Nevertheless, its main drawback is that the frames are processed individually, hence discarding temporal consistencies that are useful for getting stable and robust overlay detection and recognition system. Furthermore, processing videos on a frame-by-frame basis introduces a significant computational overhead in the context of overlay extraction - the exact problem we address in this paper. This is mainly due to the fact that the goal of overlay extraction is to output a set of phrases or sentences that do not have a significant overlap between each other, {\it i.e.} can be read as a single block of text spread across several scenes. In our approach, we tackle this problem using additional post-processing step that focuses on text rectification and proves its effectiveness through a set of qualitative results.

The problem of video overlay extraction is also tackled in~\cite{DBLP:journals/corr/KannaoG16}, where Kannao and Guha propose to detect entire lines of text instead of single words. To decrease the computational cost of detection and recognition, they use temporal tracking across multiple frames. 
For the recognition, they train the Tesseract OCR model~\cite{Smith07} with synthetically generated images. Inspired by this approach, we also generate part of our training data synthetically, however, we use the resulting dataset to improve the performance of several of our system's modules and not a Tesseract engine. Furthermore, contrary to the results presented in~\cite{DBLP:journals/corr/KannaoG16}, our text recognition engine that relies on a convolutional recurrent neural network architecture~\cite{DBLP:journals/corr/ShiBY15} significantly outperforms the competing methods, including the baseline Tesseract method. 

\section{Overlay extraction system}
\label{sec:system}
\begin{figure}[h]
\centering
\includegraphics[width=0.75\textwidth]{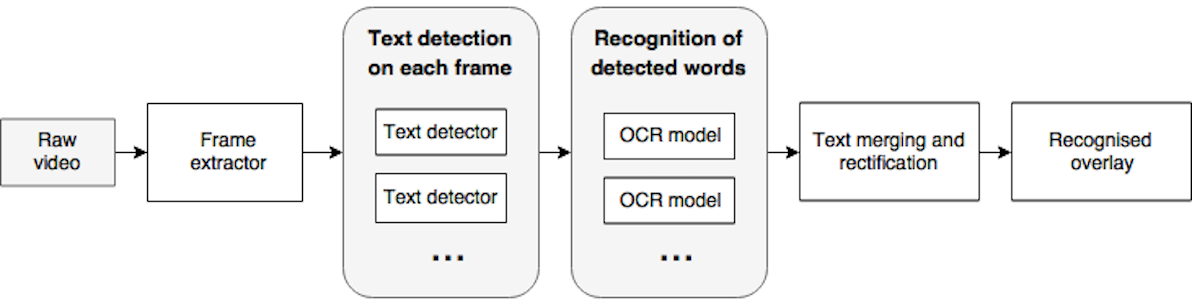}
\caption{Architecture of the system for textural overlays extraction.}
\label{fig:system}
\end{figure}

\begin{figure}[t!]
\centering
\includegraphics[width=0.48\textwidth]{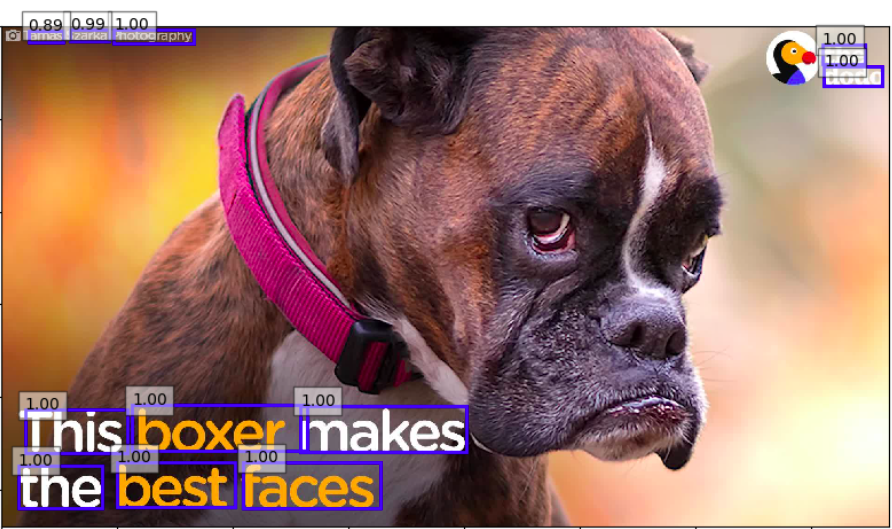}
\includegraphics[width=0.48\textwidth]{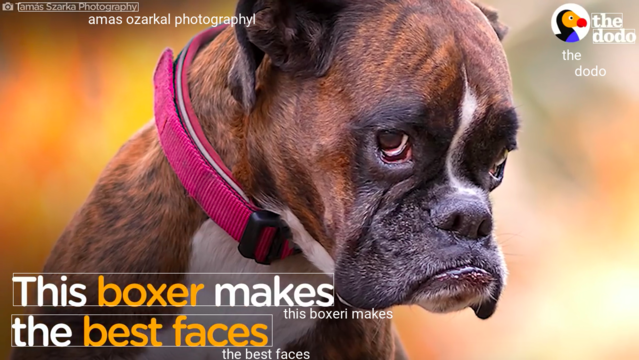}
\caption{Results of text detection \textbf{(left)} and text recognition \textbf{(right)} modules used in the proposed framework.}
\label{fig:detection_recognition}
\end{figure}

\vspace{-0.3cm} 
In this section, we present our system whose goal is to extract complete sequences of text split across several frames of a social media video. An overview of the system is also shown in Fig.~\ref{fig:system}. The proposed solution comprises several components, starting from the frame extractor through the text detector to the text recognizer and rectifier. Below, we outline the main features of those components along with the method for generating a synthetic dataset used to improve the performance of text recognition model. We conclude this section with a description of post-processing step that allows us to avoid redundancies in the overlays returned by our system. We present sample results in Fig. ~\ref{fig:detection_recognition}. 

\vspace{-0.6cm}
\subsection{Frames extractor}\label{ssec:frames-extractor}
\vspace{-0.2cm} 
The goal of this component of our system is the extraction of all frames containing unique overlays. Extracting too few frames leads to an information loss and extracting unnecessarily too many frames with overlapping overlays increases the processing time. Our frames extraction step is therefore an essential part of the whole system.

We use the functionality provided as a part of the ffmpeg codec\footnote{\url{https://www.ffmpeg.org/}} as a frame extractor. More specifically, we input a video and extract intra-coded frames, the so-called I-frames, used by the codec as benchmark frames. According to the codec specifications, I-frames are stored as complete images, in contrary to the P and B predictive frames which are encoded only through differences with respect to the benchmark I-frames. Although, there may be some cases, where the overlay text is visible only through the encoded P-frames, our preliminary results indicated that using only I-frames in those cases does not lead to a significant reduction in the information conveyed in the video.

Several alternative approaches to the problem of informative frame extraction exist. Since the overlays are typically changed when the video shot changes, selecting the last frame from every scene can be a viable solution. Unfortunately, a significant portion of our database videos consists of only one scene with multiple overlays, which reduces the applicability of this method in our use case. Another approach for frame extraction relies on a more complex method for highlight extraction based on neural network architectures~\cite{Yang15}. Our initial experiments indicated, however, that this approach is too computationally expensive and therefore reduce the usability of the entire system. We therefore rely on our frame extraction on the ffmpeg codec which provides an efficient and effective method for selecting important video frames.

\vspace{-0.4cm}
\subsection{Text detection}
\label{sec:detection}
\vspace{-0.2cm}
Our text detection component uses the TextBoxes method~\cite{Liao16} based on an end-to-end trainable Single Shot Detector (SSD) \cite{DBLP:journals/corr/LiuAESR15}. Multiple layers of the network return coordinates of word bounding boxes along with a prediction score of text presence. Then, a non-maximum suppression algorithm is used to obtain optimum bounding box coordinates for each word. The publicly available implementation\footnote{\url{https://github.com/MhLiao/TextBoxes}} we use detects only horizontal text. Therefore the text blocks that are less likely to be part of the typically horizontal overlays are automatically filtered out. In general, vertical texts are rarely used as overlays in social media videos and this is confirmed within our evaluation dataset. Modifying the solution to also detect vertical text blocks can therefore lead to a higher rate of misclassifications and ultimately reduced accuracy of our system. 
\vspace{-0.3cm}
\subsection{Text recognition}
\vspace{-0.2cm}
Our method for text recognition is based on the Convolutional Recurrent Neural Network (CRNN) model~\cite{DBLP:journals/corr/ShiBY15}. We use the architecture with seven convolutional layers followed by two Bidirectional LSTM layers. Probability of a sequence is given by a Connectionist Temporal Classification layer~\cite{Graves:2006:CTC:1143844.1143891}. 
As an input, the model takes an image displaying a single word. The image must be scaled to a fixed height while the width of the image can vary. 
Sequences that are input to recurrent layers are generated by concatenating columns of feature maps produced by convolutional layers.

Although the text detection module based on the CRNN performs well in general scenarios, its performance can be further improved by adjusting the training dataset to the application scenario. In our case, the goal is to recognize textual overlays presented in social media videos. Those videos often contain text blocks with special characters and are frequently displayed in challenging conditions (various backgrounds, font colors and sizes, etc.). To address those challenges, we propose to improve the recognition model based on the CRNN by fine-tuning the network on a synthetically generated dataset. Below, we outline the details of a dataset generation procedure which, as shown in Sec.~\ref{sec:results} leads to significant performance improvements.

\subsubsection{Generation of a synthetic dataset}
\vspace{-0.2cm}
As shown in \cite{Jaderberg14c}, training a text recognition model with synthetically generated datasets can improve its results. Furthermore, due to a specific application of our system for text recognition in social media videos, existing datasets typically used for training text recognition models may not be sufficient, as they mostly contain natural images, very much different from those published in social media. 
Therefore we propose to synthetically generate a dataset that can simulate the conditions observed in social media, such as diversified background of text blocks and various fonts and colors of the text displayed in the images. The generation of a synthetic dataset can be split into the following steps:

\textbf{Text.} We prepare transcripts of overlays from over 100 social media videos collected from several Facebook profiles and a list of 5000 most frequent words in the Corpus of Contemporary American English~\cite{COCA} 
to create a set of unique single words for rendering on images. This dataset was augmented with digits and special characters, such as hyphens, commas and question marks. This augmentation is especially important, since the original CRNN model was trained on a dataset of alphanumerical characters only and its performance is significantly decreased on a dataset of social media videos, as shown in Tab.~\ref{tab:models-metrics}. 


\textbf{Background images.} To increase the diversity of the synthetically generated dataset, we superimpose text blocks over various backgrounds. To increase the diversity of those backgrounds, we use 50 frames from randomly selected videos and manually extract regions without blocks of text. We ensure that the extracted regions represented a wide range of used colors, intensity values and texture types. 
We also extract regions whose dimensions are large enough that we can randomly crop them to increase the pool of potential background images. 

\textbf{Fonts.} We collect 71 fonts out of 30 font-families 
with Calibre font being the most popular one. The other fonts are picked to mimic the distribution of similar fonts in social media based on general guidances for editors. We present full list of font-families used below.
\begin{multicols}{3}
\begin{enumerate}
	\item{Alegreya}
    \item{Aleo}
    \item{AnonymousPro}
    \item{Archivo}
    \item{Arvo}
    \item{BioRhyme}
    \item{Bitter}
    \item{Cabin}
    \item{Calibre}
    \item{Cardo}
    \item{Chivo}
    \item{Cormorant}
    \item{CrimsonText}
    \item{Dosis}
    \item{Helvetica}
    \item{Karla}
    \item{Libre}
    \item{Lora}
    \item{Merriweather}
    \item{Montserrat}
    \item{Neuton}
    \item{OldStandard}
    \item{OpenSans}
    \item{PlayfairDisplay}
    \item{Poppins}
    \item{Raleway}
    \item{Roboto}
    \item{SourceSans}
    \item{SpaceMono}
    \item{Spectral}
\end{enumerate}
\end{multicols}

\begin{figure}[h!]
\centering
\includegraphics[width=0.3\textwidth]{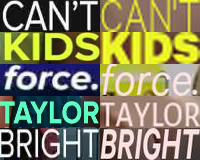}
\caption{Comparison of sample images extracted from real videos \textbf{(left column)} and synthetic images generated with the same text block~\textbf{(right column)}.}
\label{fig:comparison}
\end{figure}

\textbf{Random sampling.} For each word in our dataset, we generate 100 samples by selecting random font, size and color. Then, based on the size of the text, we crop randomly selected background image and superimpose the text on the cropped image. 
All images are resized to 100x32px with anti-aliasing and saved in jpeg format. Fig.~\ref{fig:comparison} shows a comparison between real and synthetically generated frames with text.

\subsubsection{Fine-tuning}
\vspace{-0.2cm}
We use the synthetically generated dataset to fine-tune our CRNN model. We experiment with three different variants of the CRNN tuning procedure:
\begin{enumerate}
\item  We modify the dimensions of the last LSTM layer to adapt it to the extended set of characters. Only the last LSTM layer is initialized with random weights and the parameters of all other layers are frozen.
\item We modify output dimensions of the last LSTM layer but the weights of both LSTM layers are initialized with random weights, while the other parameters are frozen.
\item We initially load weights from pretrained model and change output dimensions of the last LSTM layer. All network parameters are updated during training.
\end{enumerate}
The comparison of the results obtained with different variants is shown in Sec.~\ref{sec:results}.

\vspace{-1.0cm}
\subsection{Text merging and rectification}
\vspace{-0.2cm}
Although our frame extraction component is fairly robust, it does not prevent the text extracted from text overlays to overlap across consecutive frames. 
We propose novel yet simple method for filtering out such overlaps. We can assume that components of a single overlay appear gradually and last until the end of a scene. Therefore, the version containing the greatest number of characters can be considered the final one. We sort the extracted overlays by the time of their appearance in a reverse order. We then compare consecutive texts of overlays using normalized Levenshtein distance. If the result is below a given threshold we consider that overlays are overlapping and disregard the one with fewer characters. 
We also use an autocorrection toolkit\footnote{\url{https://github.com/phatpiglet/autocorrect/}} to further improve the results of the OCR model. 

\vspace{-0.3cm}
\section{Evaluation}
\label{sec:results}
\vspace{-0.3cm} 
In this section, we present the results of the evaluation of our method on a benchmark dataset. We first present the evaluation dataset along with the evaluation metrics. We then show the experimental results obtained for different pre-processing steps. Finally, we show the comparison of the results obtained with our method against the results of the system based on Tesseract of CRNN models.

\subsection{Dataset}
\vspace{-0.2cm}
To measure the accuracy of the OCR component of our system we extract frames from 100 videos published on Facebook between June 2017 and January 2018 on NowThisNews, NowThisPolitics, NowThisHer, thedodosite and SeekerMedia channels using ffmpeg codec, as described in section ~\ref{ssec:frames-extractor}. 
Then, using the method presented in section~\ref{sec:detection}, we detect and extract single word images from a random subset of frames. We discard images with less than 20px height. We also exclude images with less than 3 characters as well as images that are part of a media brand logo as they would introduce overlaps in our test set. We randomly select 1000 of the remaining images and manually annotate them to use as the final testing set. To measure end-to-end performance of the OCR and the text detection components, we annotate 100 randomly selected frames with 1128 words displayed in total. For each frame, we mark the location of the text and we transcribe all the words shown in the frame. The set of videos we have used to extract those frames was separate from the set that we used to select the list of words for generating synthetic images. We have not explicitly excluded repetitions of other words. We assume that random selection of frames and words taken from them is enough to prevent including two identical images in our testing set.  

\subsection{Evaluation metrics}
\vspace{-0.2cm}
To evaluate our system and compare it with the baseline, we follow the evaluation protocol of~\cite{BORNDIGITAL6065555}, and compute several metrics: average precision, recall and f1 score of the system output. 
We also compare targets with the predictions using similarity metric based on normalized Levenshtein distance~\cite{Levenshtein}. All metrics are calculated on a word level. Similar metric was used in ~\cite{LundgvistThesis} to evaluate OCR accuracy on distorted images which also may be the case in our task due to the frame extraction process. 

The metrics are computed according to the following formulas:
\begin{align*}
recall = \dfrac{|labels \cap predictions|}{|labels|} && precision = \dfrac{|labels \cap predictions|}{|predictions|}
\end{align*}
\begin{align*}
f1 = 2 \cdot \dfrac{recall \cdot precision}{recall + precision} && similarity = 1 - \dfrac{Levenshtein(labels, predictions)}{max(|labels|,|predictions|)} 
\end{align*}



\subsection{Preprocessing methods}
\vspace{-0.2cm}
The detection and recognition modules of our system expect grayscale images as their input and we evaluate several preprocessing methods that aim to improve the quality of grayscale images text recognition. 
To that end, we test the following preprocessing methods with a pretrained CRNN model~\cite{DBLP:journals/corr/ShiBY15} and the Tesseract OCR Engine~\cite{Smith:2007:OTO:1304596.1304846}:
\begin{itemize} 
\item No preprocessing: raw, grayscale images are input to the recognition component.
\item Otsu's binarization: binarization method based on dynamic thresholding. We use OpenCV\footnote{\url{http://www.opencv.org}} implementation.
\item Gaussian blurring with $5 \times 5$px kernel followed by Otsu's binarization: Additional blurring step can potentially increase the robustness of the system.
\item Gaussian blurring with Otsu's binarization and opening: by adding the morphological openning observation, we expect to reduce the amount of noise in the images.
\item Max-RGB filter: we flat the color channel space by selecting a maximum pixel value from each channel and using it as the output image pixel. This preprocessing method is based on the assumption that text and background have different color and using this filter should lead to an improved contrast of the image.
\end{itemize}

\begin{table}[t!]
\def\arraystretch{1.2}
\setlength{\tabcolsep}{0.5em}
  \begin{center}
  	\caption{Word level recognition accuracy for the Tesseract OCR engine ~\cite{Smith:2007:OTO:1304596.1304846} and pretrained CRNN model \cite{DBLP:journals/corr/ShiBY15} when given preprocessing method was applied.}
    \vspace{0.2cm}
  	\begin{tabular}{l|c|r}
    \textbf{Preprocessing method} & \textbf{Tesseract} & \textbf{CRNN model}\\
     \hline
      None& 57.8\% & 75.4\%\\ \hline
      Gaussian blur + Otsu & 52\% & 65.9\%\\
      Gaussian blur + Otsu + opening & 57.2\% & 
      67.4\%\\
      Otsu & 57.8\% & 69.1\%\\
      \textbf{Max RGB} & 56.3\% & \textbf{75.7\%}\\
    \end{tabular}
	\label{tab:tesseract-crnn}

  \end{center}
\end{table}
\vspace{-0.3cm}
Tab.~\ref{tab:tesseract-crnn} shows the results of the experiments with preprocessing methods. For the Tesseract OCR the best pre-processing method is Otsu's binarization, yet identical result was obtained without the preprocessing. However, for the CRNN model, which we use in our system in practice, the best results are obtained when using max-RGB filtering. Nevertheless, the performance improvement achieved by the best preprocessing method is negligible. The conclusion of this experiment is that the convolutional layers of the CRNN module are able to learn optimal transformations to increase the system performance and therefore fully substitute preprocessing steps. One can also see that the neural network based model significantly outperforms the traditional Tesseract OCR system. 

\subsection{Results}
\vspace{-0.3cm}
Accuracy tests performed with the original model and its fine-tuned versions presented in the Table \ref{tab:models-metrics} show improvement for cases where only parameters for LSTM layers were updated during training. It means that the generated set may be too small for training the entire network without overfitting. At the same time updating parameters of both LSTM layers turn out to be better than modifying parameters of only the single last layer. It shows that features encoded by the penultimate LSTM layer are not generic enough and that our synthetic training set is sufficient to learn new features specific to the task. The best model allowed to reduce the word recognition error by 20\%.

Evaluation of text detection and recognition presented in Table \ref{tab:models-metrics} shows that all fine-tuned models perform better than the original version for this specific task. Using the fine-tuned CRNN model leads to a 20\% increase of precision, recall, F1 score and similarity metrics compared to a generic CRNN model.

End-to-end results show that the text detection component plays an important role in the system. Imperfect detection lowers the quality of CRNN input which translates into a decrease in recognition accuracy. However, from a practical point of view the system can be already used to extract meaningful information from social media videos. Overlays can be further processed using presented rectification methods.  

\begin{table}[t!]
 \def\arraystretch{1.2}
 \setlength{\tabcolsep}{0.6em}
  \begin{center}
  \caption{Performance comparison of baseline, original and fine-tuned models.}
  \vspace{0.2cm}
  \begin{tabular}{p{3.3cm}|c|c|c|c|c}
    & {\it Recognition} & \multicolumn{4}{c}{ {\it Detection with recognition}} \\
  \hline
  \textbf{Model} & \textbf{Accuracy} & \textbf{Precision} & \textbf{Recall} & \textbf{F1} & \textbf{Similarity} \\
  \hline
	Tesseract~\cite{Smith:2007:OTO:1304596.1304846} &  57.8\% & 0.284 & 0.266 & 0.274 & 0.42\\
	CRNN~\cite{DBLP:journals/corr/ShiBY15} & 75.7\% & 0.368 & 0.343 & 0.352 & 0.52\\
	Fine-tuned CRNN (all parameters) & 74\% & 0.40 & 0.375 & 0.386 & 0.59\\
	Fine-tuned CRNN (last LSTM layer) & 76.6\%  & 0.406 & 0.378 & 0.389 & 0.59\\
	\textbf{Fine-tuned CRNN (both LSTM layers)} & \textbf{80.1\%} & \textbf{0.45} & \textbf{0.42} & \textbf{0.432} & \textbf{0.62}\\
  \end{tabular}
  \label{tab:models-metrics}
  \end{center}
 \end{table}

\begin{figure}[t!]
    \centering
    \subfloat[Tesseract]{{\includegraphics[width=0.75\textwidth]{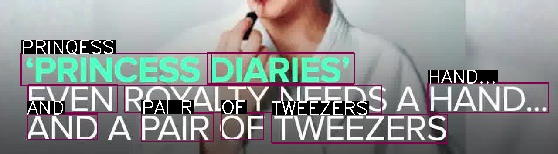} }}%

    \subfloat[Original CRNN model ~\cite{DBLP:journals/corr/ShiBY15}]{{\includegraphics[width=0.75\textwidth]{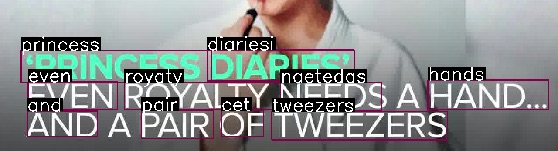} }}%

        \subfloat[Fine-tuned CRNN model]{{\includegraphics[width=0.75\textwidth]{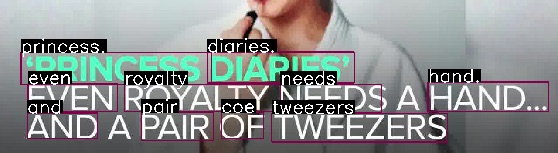} }}%

        
    \caption{Results of text recognition using different models.}%
    \vspace{-0.5cm}
    \label{fig:recognition-example}%
\end{figure}

To further evaluate different models used for text recognition, we visualize the outputs of various methods on a sample video frame with the overlay. The results are shown in Fig.~\ref{fig:recognition-example}. Those qualitative results confirm the numerical evaluation performed above - our fine-tuned CRNN model provides the most accurate transcription of the overlay. 

\vspace{-0.4cm}
\section{Conclusions}
\label{sec:conclusions}
\vspace{-0.2cm}
In this paper, we presented a comprehensive system for video overlay text extraction that comprises several components: keyframe extraction, text detection, recognition and rectification. The system is specifically designed and evaluated in the context of social media videos where textual overlays appear in particularly challenging conditions. Using synthetically generated dataset allowed us to reduce recognition error of our neural network-based text recognition model by over 20\%. Overall, the proposed system provides an effective and robust method for video overlay extraction. It has been successfully implemented and integrated into a complex social media video analysis engine and is actively used as part of many services, including a content classifier and a retention analytics engine. 

\vspace{-0.3cm}
\subsection*{Acknowledgments}
\vspace{-0.2cm}
{\small
This work was partially funded by the Dean's Grant nr II/2017/GD/1 of the Faculty of Electronics and Information Technology at Warsaw University of Technology.}
\vspace{-0.2cm}

\bibliographystyle{unsrt}

\end{document}